\documentclass[conference,a4paper]{IEEEtran}
\usepackage{eso-pic}
\usepackage{graphicx,booktabs,amsmath,multirow,array}
\usepackage{newtxtext,newtxmath}
\usepackage{tikz}
\usepackage[colorlinks,allcolors=black,urlcolor=blue]{hyperref}

\newcommand{\LeakMaxChange}{0.005}

\newcommand{\SignModel}{0.016}

\newcommand{\NGraphSig}{4}

\newcommand{\NSigDgtFm}{5}

\newcommand{\NAheadDescFm}{7}
\newcommand{\NSigDescFm}{3}

\newcommand{\NDescHurtReg}{2}
\newcommand{\NDescRaiseCls}{4}

\newcommand{\NBeatDgtFm}{6}
\newcommand{\NBeatDgtRf}{3}

\newcommand{\NDgtRepro}{3}
\newcommand{\NFmBeatsPubUnrepro}{4}

\newcommand{\NFpRegCells}{9}
\newcommand{\NFpRegDgtAhead}{7}

\newcommand{\PtFmMaccsInputRefEsol}{\ensuremath{0.419}}

\newcommand{\PtFmMaccsInputCmpEsol}{\ensuremath{0.415}}

\newcommand{\PtFmMaccsModelStartEsol}{\ensuremath{0.055}}

\newcommand{\PtFmMaccsModelRdkitEsol}{\ensuremath{0.051}}

\newcommand{\PtFmMaccsInputMinusModelEsol}{\ensuremath{0.364}}
\newcommand{\PtFmMaccsInputMinusModelEsolLo}{\ensuremath{0.210}}
\newcommand{\PtFmMaccsInputMinusModelEsolHi}{\ensuremath{0.515}}

\newcommand{\PtFmMaccsInputRefFreesolv}{\ensuremath{-0.110}}

\newcommand{\PtFmMaccsInputCmpFreesolv}{\ensuremath{0.196}}

\newcommand{\PtFmMaccsInputMinusModelFreesolv}{\ensuremath{-0.321}}
\newcommand{\PtFmMaccsInputMinusModelFreesolvLo}{\ensuremath{-0.815}}
\newcommand{\PtFmMaccsInputMinusModelFreesolvHi}{\ensuremath{0.108}}

\newcommand{\PtPfnEcfpInputMinusModelFreesolv}{\ensuremath{0.804}}
\newcommand{\PtPfnEcfpInputMinusModelFreesolvLo}{\ensuremath{0.295}}
\newcommand{\PtPfnEcfpInputMinusModelFreesolvHi}{\ensuremath{1.309}}

\newcommand{\PtFmMaccsModelRdkitLipo}{\ensuremath{0.100}}
\newcommand{\PtFmMaccsModelRdkitLipoLo}{\ensuremath{0.053}}
\newcommand{\PtFmMaccsModelRdkitLipoHi}{\ensuremath{0.145}}

\newcommand{\PtFmMaccsInputMinusModelLipo}{\ensuremath{0.101}}
\newcommand{\PtFmMaccsInputMinusModelLipoLo}{\ensuremath{0.043}}
\newcommand{\PtFmMaccsInputMinusModelLipoHi}{\ensuremath{0.158}}

\newcommand{\PtFmMaccsInteractionLipo}{\ensuremath{0.090}}
\newcommand{\PtFmMaccsInteractionLipoLo}{\ensuremath{0.038}}
\newcommand{\PtFmMaccsInteractionLipoHi}{\ensuremath{0.141}}

\newcommand{\PtPfnEcfpInteractionLipo}{\ensuremath{0.153}}
\newcommand{\PtPfnEcfpInteractionLipoLo}{\ensuremath{0.102}}
\newcommand{\PtPfnEcfpInteractionLipoHi}{\ensuremath{0.202}}

\newcommand{\StrictMinMargin}{\ensuremath{0.004}}
\newcommand{\StrictMinModel}{TabPFN}
\newcommand{\StrictMinStart}{ECFP4}
\newcommand{\StrictMinDataset}{Lipophilicity}

\newcommand{\ClsSigModel}{CatBoost}
\newcommand{\ClsSigStart}{ECFP4}
\newcommand{\ClsSigDataset}{SIDER}
\newcommand{\ClsSigEst}{\ensuremath{0.054}}

\newcommand{\ExtNEsol}{6{,}954}

\newcommand{\ExtFmInputAvgEsol}{\ensuremath{0.337}}
\newcommand{\ExtFmInputAvgEsolLo}{\ensuremath{0.238}}
\newcommand{\ExtFmInputAvgEsolHi}{\ensuremath{0.414}}

\newcommand{\ExtFmModelAvgEsol}{\ensuremath{0.064}}
\newcommand{\ExtFmModelAvgEsolLo}{\ensuremath{0.021}}
\newcommand{\ExtFmModelAvgEsolHi}{\ensuremath{0.099}}

\newcommand{\ExtFmModelRdkitEsol}{\ensuremath{0.079}}
\newcommand{\ExtFmModelRdkitEsolLo}{\ensuremath{0.019}}
\newcommand{\ExtFmModelRdkitEsolHi}{\ensuremath{0.129}}

\newcommand{\ExtNLipo}{2{,}668}

\newcommand{\ExtFmInputAvgLipo}{\ensuremath{0.195}}
\newcommand{\ExtFmInputAvgLipoLo}{\ensuremath{0.161}}
\newcommand{\ExtFmInputAvgLipoHi}{\ensuremath{0.229}}

\newcommand{\ExtFmModelAvgLipo}{\ensuremath{0.068}}
\newcommand{\ExtFmModelAvgLipoLo}{\ensuremath{0.043}}
\newcommand{\ExtFmModelAvgLipoHi}{\ensuremath{0.093}}

\newcommand{\ExtFmModelRdkitLipo}{\ensuremath{0.092}}
\newcommand{\ExtFmModelRdkitLipoLo}{\ensuremath{0.061}}
\newcommand{\ExtFmModelRdkitLipoHi}{\ensuremath{0.122}}

\newcommand{\XsqOther}{7}
\newcommand{\XsqMain}{5}

\newcommand{\XsqNewRdkit}{4}

\newcommand{\XsqRtfN}{5}
\newcommand{\XsqRtfEsolLo}{\ensuremath{0.270}}
\newcommand{\XsqRtfEsolHi}{\ensuremath{0.547}}
\newcommand{\XsqRtfLipoLo}{\ensuremath{0.082}}
\newcommand{\XsqRtfLipoHi}{\ensuremath{0.159}}
\newcommand{\XsqRtfLipoStrict}{3}

\newcommand{\XsqMgN}{6}
\newcommand{\XsqMatchedEsolFull}{0.63}
\newcommand{\XsqMatchedEsolFolded}{0.61}

\newcommand{\XsqMgEsolLo}{\ensuremath{0.552}}
\newcommand{\XsqMgEsolHi}{\ensuremath{0.704}}
\newcommand{\XsqMatchedLipoFull}{0.15}
\newcommand{\XsqMatchedLipoFolded}{0.12}

\newcommand{\XsqMgLipoLo}{\ensuremath{0.089}}
\newcommand{\XsqMgLipoHi}{\ensuremath{0.247}}

\newcommand{\ExtRtfSolLo}{\ensuremath{0.128}}
\newcommand{\ExtRtfSolHi}{\ensuremath{0.285}}

\newcommand{\ExtRtfLipoLo}{\ensuremath{0.050}}
\newcommand{\ExtRtfLipoHi}{\ensuremath{0.127}}
\newcommand{\ExtRtfLipoStrict}{1}

\newcommand{\SimNCells}{15}
\newcommand{\SimInputAll}{12}

\newcommand{\SimInputFar}{12}

\newcommand{\SimRtfNCells}{15}
\newcommand{\SimRtfInputAll}{10}

\newcommand{\SimRtfInputFar}{10}

\begin{document}
\bstctlcite{IEEEexample:BSTcontrol}
\title{Representation and Model Contributions on Small Molecular Benchmarks: Controlled Comparisons and External Evaluation}
\author{\IEEEauthorblockN{Nora A. M. A. Talkhan}
\IEEEauthorblockA{\textit{Faculty of Computing and}\\
\textit{Artificial Intelligence}\\
\textit{Cairo University}, Giza, Egypt\\
n.abdelhameed@fci-cu.edu.eg}
\and[\hspace{20pt}]
\IEEEauthorblockN{Ibrahim Abdelbaky}
\IEEEauthorblockA{\textit{Faculty of Computers and Information Technology}\\
\textit{Future University}, Cairo, Egypt\\
\textit{Faculty of Computers and Artificial Intelligence}\\
\textit{Benha University}, Banha, Egypt\\
Ibrahim.Abdelbaky@fue.edu.eg}
\and[\hspace{20pt}]
\IEEEauthorblockN{Fatma A. Omara}
\IEEEauthorblockA{\textit{Faculty of Computing and}\\
\textit{Artificial Intelligence}\\
\textit{Cairo University}, Giza, Egypt\\
f.omara@fci-cu.edu.eg}}
\maketitle
\AddToShipoutPictureFG*{\AtPageLowerLeft{\put(\LenToUnit{\dimexpr(\paperwidth-\textwidth)/2\relax},\LenToUnit{2.2cm}){\parbox[b]{\textwidth}{\scriptsize\noindent \copyright~2026 IEEE. Personal use of this material is permitted. Permission from IEEE must be obtained for all other uses, in any current or future media, including reprinting/republishing this material for advertising or promotional purposes, creating new collective works, for resale or redistribution to servers or lists, or reuse of any copyrighted component of this work in other works.}}}}

\begin{abstract}
Improvements in molecular property prediction are usually attributed to the model, although part of
the gain may come from the molecular representation. New models are commonly benchmarked against a
random forest (RF) trained on a molecular fingerprint, a comparison that changes the model and the
representation at the same time. We separate the two contributions with a two-by-two factorial
design in which the RF and a second model (CatBoost or the tabular foundation models TabPFN and
TabFM) are each trained on a fingerprint and on RDKit descriptors, and each effect is tested with a
paired scaffold bootstrap. Seven small MoleculeNet datasets (3 regression, 4 classification) are
evaluated on the published split of a recent graph transformer (DGT), on 10 additional scaffold
splits and on external datasets. For solubility and lipophilicity, the representation effect
exceeds the model effect in every main factorial design, in all three evaluations. For
hydration free energy, the result depends on the starting fingerprint: from ECFP4 the
representation effect is larger, and from MACCS neither effect is significantly larger. On the
classification datasets, only one main factorial design has a significant difference between the
two effects. Although the model effect is smaller on solubility
and lipophilicity, TabFM outperforms the RF on all 7 datasets when both use RDKit descriptors. With
RDKit descriptors in place of a fingerprint, the RF baseline reported for DGT surpasses the
transformer's published solubility score. We recommend that model comparisons evaluate classical
baselines on more than one representation and report the representation and model effects
separately.
\end{abstract}

\renewcommand{\IEEEkeywordsname}{Keywords}

\section{Introduction}

Most molecular property datasets are small. The MoleculeNet datasets used to rank new
models~\cite{wu_moleculenet_2018} often hold one or two thousand molecules, and some fewer than
one thousand. Yet new models are still presented as the top performers on these datasets,
typically against a classical baseline built on ECFP fingerprints. For example, the DGT graph
transformer~\cite{zhang_enhancing_2026} reports state-of-the-art results on 10 MoleculeNet
datasets while using only a random forest on ECFP4 as its classical baseline. Similarly, a
recent graph attention network study~\cite{waleed_statistical_2026} compares graph models with
classical methods on ECFP fingerprints, but does not report any tuning for those baselines.

A second example comes from tabular foundation models. TabPFN~\cite{hollmann_accurate_2025} and
Google's TabFM~\cite{kong_tabfm_2026} are pretrained on synthetic tables and predict a new
table by reading its labeled rows as context, without updating their weights. Using RDKit
descriptors for molecular tasks, these models match or exceed boosted trees, and in some
studies even trained graph
models~\cite{ben_hicham_tabular_2026,matyushin_-context_2026,chen_tabpfn_2026,guan_can_2026,zhao_revisiting_2026}.

In both cases the comparison changes two things at once: the model and its molecular
representation, that is, the features the model is given to read. A graph model reads the
molecular graph, a tabular model reads a table of descriptors, and the usual baseline reads a
fingerprint. A tabular model can be given another representation while the model stays the
same, so the two can be measured separately. When the new model wins, the improvement may come
from the model, from the representation or from both. If the representation is responsible for
most of the improvement, then the weak baseline is better strengthened by giving it a more
informative representation. In that case, the modeling advance is smaller than it seems.

Earlier studies have found that the choice of molecular representation often matters
more than the choice of
model~\cite{stepisnik_comprehensive_2021,orosz_comparison_2022,van_tilborg_exposing_2022,deng_systematic_2023}.
Kamuntavi\v{c}ius et al.~\cite{kamuntavicius_benchmarking_2025} crossed representations with models
and used hypothesis tests to choose the best representation for each dataset.
Most of these studies use different datasets and splits from those behind the best
published scores. None of them reports the representation effect, the model effect and their interaction,
each with a confidence interval. We measure them for the RF and 3 other tabular models on the test sets
behind DGT's published scores, with a statistical test for each dataset. Our contributions are:
\begin{itemize}
\item \emph{A $2\times2$ factorial design applied to model claims.} The factorial design is a standard
method~\cite{montgomery_design_2019}, and we apply it to a claimed improvement. We cross the two models
with the two representations and measure the representation effect, the model effect and their
interaction, each with a 95\% confidence interval, and test whether the representation is the larger
effect in either order (Section~\ref{sec:effects}).
\item \emph{A controlled comparison of models and representations on the published test sets.} We run 4 tabular
models on 4 representations, next to 2 graph models, on 7 MoleculeNet datasets, using the split
behind DGT's published scores and 10 seeded scaffold splits, with a paired test for every
dataset (Sections~\ref{sec:res-representation} and~\ref{sec:res-limits}).
\item \emph{A check of the published baseline.} We replace ECFP4 with RDKit descriptors in DGT's RF
baseline and compare it with DGT's published scores, and we rerun DGT itself
(Section~\ref{sec:res-published}).
\item \emph{External datasets.} The same trained models are evaluated on solubility and
lipophilicity data from other sources (Section~\ref{sec:res-external}).
\end{itemize}
The code is available at \url{https://github.com/NoraRo/representation-vs-model}.

\section{Related Work}

\emph{Representation or model.} Jiang et
al.~\cite{jiang_could_2021} found that descriptor-based models performed on average as well as or
better than graph neural networks on 11 datasets. Van Tilborg et al.~\cite{van_tilborg_exposing_2022}
found that the descriptor explained most of the performance differences on 30 bioactivity datasets,
with ECFP giving the best results, and Orosz et al.~\cite{orosz_comparison_2022} found 2D descriptors best for ADME-Tox
classification. In contrast, Yang et al.~\cite{yang_analyzing_2019} found that their
message passing network matched or beat descriptor-based models, and that adding RDKit features helped
on some datasets, such as ESOL, but not on others. Stepi\v{s}nik et
al.~\cite{stepisnik_comprehensive_2021} compared fingerprints, descriptors and learned
representations on 11 datasets and found descriptors best for physical properties.
Kamuntavi\v{c}ius et al.~\cite{kamuntavicius_benchmarking_2025} crossed many representations with
5 models on ADMET data, used hypothesis tests to choose the best representation for each
dataset, and tested models trained on one source on data from another. Deng et al.~\cite{deng_systematic_2023} found that an RF on RDKit 2D descriptors beats
pretrained models reported as state-of-the-art on most MoleculeNet datasets, and beats Morgan
fingerprints on BBBP, ESOL, FreeSolv and Lipophilicity, using 30 seeded splits rather than the
split behind published numbers. These studies compare representations and models to find the best pair, but none
reports the representation effect, the model effect and their interaction.

\emph{Tabular foundation models on molecules.} Ben Hicham et
al.~\cite{ben_hicham_tabular_2026} evaluated TabPFN and TabICL with 12 representations on 58 Polaris
and MoleculeACE tasks. The models matched or beat trained models, including Chemprop, and the
representation was a main factor in their ranking. Chen et al.~\cite{chen_tabpfn_2026} found
TabPFN as good as XGBoost on classification and better on regression. Matyushin and
Sholokhova~\cite{matyushin_-context_2026} used one representation for all models on 11
regression datasets and found that tabular foundation models won more often than tuned boosted
trees across those datasets. These studies either keep the representation fixed or rank each
model and representation pair as a whole, so none can tell the two effects apart, and none
evaluates Google's TabFM on molecules.

\emph{How MoleculeNet is scored.} Sun et
al.~\cite{sun_does_2022} reported that the same graph network scored 0.922 ROC-AUC on BBBP with a
balanced scaffold split but 0.748 with the deterministic scaffold split. Scaffold splits are also easier
than splits by similarity clusters~\cite{fooladi_evaluating_2025}, and these datasets contain
label conflicts and repeated molecules~\cite{schuh_auditing_2026}. Ash et
al.~\cite{ash_practically_2025} argued that a fair comparison needs repeated splits and a named
statistical test.

\section{Methods}\label{sec:method}

\subsection{Datasets and Splits}
We use 7 MoleculeNet datasets~\cite{wu_moleculenet_2018}: 3 regression datasets, ESOL
(aqueous solubility), FreeSolv (hydration free energy) and Lipophilicity, and 4
classification datasets, BBBP, BACE, ClinTox and SIDER. These are the 7
of DGT's 10 datasets with at most 4,200 molecules. We leave out Tox21, HIV and QM9 because
tabular foundation models read all training rows as context at prediction time, so they are built for small tables. We read the standard DeepChem files that DGT loads and drop only the rows RDKit cannot
parse. The datasets hold 642 to 4,200 molecules, and the
published split leaves 65 to 420 of them for testing (Supplementary S1).

Every split groups molecules by their Bemis--Murcko scaffold~\cite{bemis_properties_1996}, so
no scaffold is shared between training and test data. The \emph{published split} is the rule
in DGT's released code: scaffold groups, largest first, fill training up to 80\% of the data,
then validation up to 90\%, and the rest is test. So our test sets are the ones behind DGT's
published numbers. We also build 10 \emph{seeded splits} in the style of Chemprop's balanced
scaffold split~\cite{heid_chemprop_2024}, at 80:10:10, to see how much a result depends on the split.

In the published split, a few test molecules in BBBP, BACE and SIDER also appear in the training
data, as a salt or a charged form of the same molecule. We keep them, because the published scores
include them. Removing them changes no result by more than \LeakMaxChange{} (Supplementary S2).

The \emph{external datasets} check whether the regression result holds on molecules from other
sources. We use them for ESOL and Lipophilicity, where the representation effect is significantly larger on
the published split. FreeSolv has no external dataset, because we found no hydration free energy set
independent of it. For ESOL we use AqSolDBc~\cite{llompart_will_2024}, a curated aqueous solubility
set. For Lipophilicity we use the T-data of RTlogD~\cite{wang_logd74_2023}, which holds
logD$_{7.4}$ values added to ChEMBL between releases 29 and 32. We remove every molecule that also
occurs in the MoleculeNet dataset, and a few unusable ones (Supplementary S1), which leaves
\ExtNEsol{} solubility and \ExtNLipo{} lipophilicity molecules. The models trained on MoleculeNet
are applied to them without refitting or tuning.

New molecules also exist inside MoleculeNet. For each test molecule, we find its most similar
training or validation molecule by ECFP4 Tanimoto similarity, and cut the test set into a
\emph{low-similarity} half and a \emph{high-similarity} half. We then compute every factorial design
again within each half and check whether each representation step is still larger than each model
step (Supplementary S5).

\subsection{Representations and Models}
We use the 217 RDKit 2D descriptors, called \emph{rdkit2d} in the tables and equations, which
cover counts, charges, surface areas, logP and similar values. We also use the ECFP4 Morgan
fingerprint~\cite{rogers_extended-connectivity_2010} with 2,048 bits and the 166 MACCS structural
keys~\cite{durantReoptimizationMDLKeys2002}. Because TabFM does not run on ECFP4 on all 7 datasets
(Section~\ref{sec:limitations}), we also use the same Morgan fingerprint (radius 2) folded to 512 bits,
called \emph{morgan512} in the tables. All 4 are computed with the RDKit library, version
2026.03.5. All tabular models read the same feature
matrix.

The tabular models are the RF~\cite{breiman_random_2001}, CatBoost~\cite{prokhorenkova_catboost_2018}
and 2 foundation models, TabPFN-3~\cite{hollmann_accurate_2025,grinsztajn_tabpfn-3_2026} and
TabFM~\cite{kong_tabfm_2026}. We tune the RF and CatBoost by keeping the best of 4 settings
on the validation score, and the foundation models use library defaults except for 8 ensemble
members. All tabular models use the training and
validation rows.

We also run 2 graph models. Chemprop~\cite{yang_analyzing_2019,heid_chemprop_2024} is tuned the
same way, over 4 settings of batch normalization and dropout, and DGT~\cite{zhang_enhancing_2026} uses the code and settings of its release on the
published split. We compare with DGT in two ways. Its published scores come without
predictions for each test molecule, so we compare with them without a statistical test. Our own DGT runs give a prediction for each test
molecule, so we compare with them using the same paired test as for any other model. Our runs reach the published score on \NDgtRepro{} of
the 7 datasets (Supplementary S6). We also run 3 variants as checks: TabFM's full
preset (TabFM+), Chemprop refit on training plus validation, and Chemprop with the RDKit
descriptors added after message passing. Their results are in Section~\ref{sec:res-published}
and Supplementary S6. Versions and settings are in the code repository, and run time and memory in Supplementary S7.

\subsection{Factorial Design and Effect Definitions}\label{sec:effects}
\begin{table}[!htb]
\centering
\footnotesize
\setlength{\tabcolsep}{3pt}
\caption{What each analysis answers.}
\label{tab:analyses}
\begin{tabular}{>{\raggedright\arraybackslash}p{5.85cm}>{\raggedright\arraybackslash}p{2.6cm}}
\toprule
Question & Answered by \\
\midrule
How much did each change improve the score? & Representation effect $R$, model effect $M$ \\
Which change contributed more? & $R-M$ \\
Does the model step depend on the representation? & Interaction \\
Is the representation larger in either order? & $R-M>|\text{interaction}|$ \\
Could a difference come from the choice of test molecules? & Scaffold bootstrap, 95\% interval \\
Is it still significant after testing 7 datasets? & Holm correction \\
Is it stable across splits? & 10 seeded splits \\
Does it hold on new molecules? & External datasets \\
\bottomrule
\end{tabular}
\end{table}
For a model $m$ on representation $x$, let $L(m,x)$ be the test loss, which is the RMSE for
regression and $1-\text{AUC}$ for classification. We measure each change as the loss before it
minus the loss after it, so a positive value is always an improvement.

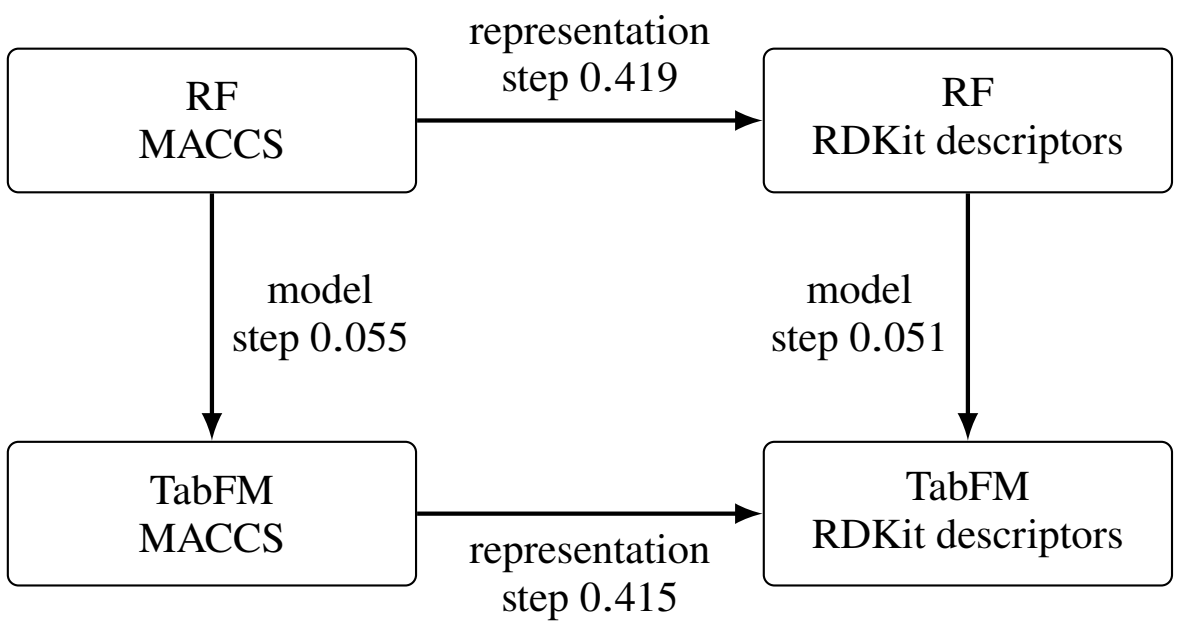
\begin{figure}[t]
\centering
\begin{tikzpicture}[box/.style={draw, rounded corners=2pt, minimum width=2.7cm,
  minimum height=0.95cm, align=center, font=\footnotesize}, lab/.style={align=center, font=\footnotesize},
  >=latex]
\node[box] (a) at (0,0) {RF\\MACCS};
\node[box] (b) at (5.0,0) {RF\\RDKit descriptors};
\node[box] (c) at (0,-2.6) {TabFM\\MACCS};
\node[box] (d) at (5.0,-2.6) {TabFM\\RDKit descriptors};
\draw[->, thick] (a) -- node[lab, above] {representation\\step \PtFmMaccsInputRefEsol} (b);
\draw[->, thick] (c) -- node[lab, below] {representation\\step \PtFmMaccsInputCmpEsol} (d);
\draw[->, thick] (a) -- node[lab, right] {model\\step \PtFmMaccsModelStartEsol} (c);
\draw[->, thick] (b) -- node[lab, left] {model\\step \PtFmMaccsModelRdkitEsol} (d);
\end{tikzpicture}
\caption{One factorial design, on the published ESOL test set. Each arrow is the RMSE improvement from one change.}
\label{fig:square}
\end{figure}

To separate the two changes (Table~\ref{tab:analyses}), we use a $2\times2$ factorial design~\cite{montgomery_design_2019}.
It runs the RF and a model $m$, each on a starting representation $x$ and on RDKit descriptors. Fig.~\ref{fig:square} shows
one such design. Its corners are the 4 model--representation
pairs. Each horizontal arrow changes only the representation (a \emph{representation step}), and
each vertical arrow changes only the model (a \emph{model step}). Going from
$(\text{RF},x)$ to $(m,\text{rdkit2d})$ takes one arrow of each kind, in either order. Both orders
reach the same total improvement, but they can split it differently, because a model step may be
larger on one representation than on the other. Since neither order is more correct, the
\emph{representation effect} $R$ is the average of the two representation arrows and the \emph{model
effect} $M$ the average of the two model arrows. Together, $R$ and $M$ add up to the total improvement.

$R-M$ says which change contributed more. It equals
\mbox{$L(m,x) - L(\text{RF},\text{rdkit2d})$}, so it is positive exactly when the RF on RDKit
descriptors beats the other model on the starting representation. The interaction says
whether the model step depends on the representation. It is the model arrow on RDKit descriptors
minus the model arrow on the starting representation, and the two representation arrows differ by
the same amount. Because a large interaction can make
averaged effects misleading~\cite[Sec.~5.1]{montgomery_design_2019}, we call the representation
\emph{the larger effect in either order} only when $R-M$ exceeds the absolute value of the interaction. In
this case, each representation arrow is larger than each model arrow.

We report \XsqMain{} \emph{main factorial designs}, all with the RF as reference, because the RF is the
classical baseline of the graph papers we compare with: TabFM, TabPFN and CatBoost from MACCS,
and TabPFN and CatBoost from ECFP4. To check that the result does not depend on the RF as
reference model or on RDKit descriptors as the final representation, we also build the \XsqOther{} other possible factorial designs
(Supplementary S3). Each effect belongs to its factorial design, so it
describes the 4 model--representation pairs in that design, not a general property of a model or a
representation.

\subsection{Statistics}
On the published split, we give each score difference between two model--representation pairs a
95\% confidence interval. We redraw the test set 10,000 times, recompute the difference each time
and take the middle 95\% of the values. Each draw takes whole scaffold groups, because molecules
sharing a scaffold are not independent. All models are scored on the same draws, so the two
effects, their difference and their interaction also each get an interval.
Each of these 4 numbers is tested on 7 datasets in each factorial design. We adjust its 7 p-values with
Holm's method~\cite{holm_simple_1979} and call it \emph{significant} only when the adjusted p-value
is below 0.05 and the interval excludes zero. To test whether an effect points the same way across
the 7 datasets, we use an exact sign test (Supplementary S1). SIDER and ClinTox are scored as
the mean AUC over their tasks, so each dataset counts once.

The seeded splits share most of their training molecules, so we report how many of the 10 favor
each side, as a measure of stability, not significance. To look at several models and representations at once
instead of one factorial design at a time, we also fit a two-way analysis of variance to the seeded scores of RF,
CatBoost and TabPFN on ECFP4, MACCS and RDKit descriptors~\cite{montgomery_design_2019}.

\section{Results}\label{sec:results}

\subsection{Representation Effect Larger Than Model Effect}\label{sec:res-representation}
\begin{table}[!htb]
\centering
\footnotesize
\setlength{\tabcolsep}{3pt}
\caption{Representation effect $-$ model effect on the published split. Positive: the representation helped more. Factorial design: (RF, start) $\to$ (model, RDKit descriptors). Bold: significant.}
\label{tab:squares}
\begin{tabular}{clrrrrr}
\toprule
 &  & \multicolumn{3}{c}{From MACCS} & \multicolumn{2}{c}{From ECFP4} \\
\cmidrule(lr){3-5} \cmidrule(lr){6-7}
 & Dataset & TabFM & TabPFN & CatBoost & TabPFN & CatBoost \\
\midrule
\multirow{3}{*}{\rotatebox[origin=c]{90}{RMSE}} & ESOL & $\mathbf{0.364}$ & $\mathbf{0.377}$ & $\mathbf{0.412}$ & $\mathbf{0.607}$ & $\mathbf{0.641}$ \\
 & FreeSolv & $-0.321$ & $-0.247$ & $-0.305$ & $\mathbf{0.804}$ & $0.919$ \\
 & Lipophilicity & $\mathbf{0.101}$ & $\mathbf{0.084}$ & $\mathbf{0.110}$ & $\mathbf{0.157}$ & $\mathbf{0.080}$ \\
\midrule
\multirow{4}{*}{\rotatebox[origin=c]{90}{AUC}} & BBBP & $-0.006$ & $0.013$ & $0.032$ & $0.039$ & $0.058$ \\
 & BACE & $-0.047$ & $-0.036$ & $-0.049$ & $-0.056$ & $-0.001$ \\
 & ClinTox & $-0.024$ & $-0.012$ & $-0.036$ & $0.162$ & $0.102$ \\
 & SIDER & $0.008$ & $0.028$ & $0.020$ & $0.039$ & $\mathbf{0.054}$ \\
\bottomrule
\end{tabular}
\end{table}

\begin{table*}[t]
\centering
\footnotesize
\caption{Scores on the published split, grouped by representation. $\scriptstyle\downarrow$\,/\,$\scriptstyle\uparrow$: lower/higher is better. Bold: best score in each column (not necessarily significant). DGT (published): mean of 4 runs, from DGT's source data.}
\label{tab:scores}
\begin{tabular}{lrrrrrrr}
\toprule
 & \multicolumn{3}{c}{Regression: RMSE $\downarrow$} & \multicolumn{4}{c}{Classification: ROC-AUC $\uparrow$} \\ \cmidrule(lr){2-4} \cmidrule(lr){5-8}
Model & ESOL & FreeSolv & Lipo & BBBP & BACE & ClinTox & SIDER \\
\midrule
\multicolumn{8}{l}{\emph{rdkit2d}} \\
\quad RF & $0.766$ & $2.256$ & $0.747$ & $0.750$ & $0.818$ & $0.890$ & $0.689$ \\
\quad CatBoost & $0.739$ & $2.032$ & $0.698$ & $0.730$ & $0.818$ & $0.914$ & $0.685$ \\
\quad TabPFN & $0.733$ & $1.767$ & $0.646$ & $\mathbf{0.766}$ & $0.848$ & $0.937$ & $0.689$ \\
\quad TabFM & $0.715$ & $\mathbf{1.739}$ & $0.647$ & $0.757$ & $0.847$ & $0.947$ & $0.695$ \\
\quad TabFM+ & $\mathbf{0.711}$ & $1.816$ & $\mathbf{0.641}$ & $0.752$ & $0.852$ & $\mathbf{0.948}$ & $\mathbf{0.699}$ \\
\midrule
\multicolumn{8}{l}{\emph{ECFP4}} \\
\quad RF & $1.450$ & $3.587$ & $0.851$ & $0.705$ & $\mathbf{0.885}$ & $0.798$ & $0.668$ \\
\quad CatBoost & $1.408$ & $3.175$ & $0.826$ & $0.692$ & $0.819$ & $0.788$ & $0.635$ \\
\quad TabPFN & $1.373$ & $3.060$ & $0.903$ & $0.711$ & $0.874$ & $0.728$ & $0.650$ \\
\midrule
\multicolumn{8}{l}{\emph{MACCS}} \\
\quad RF & $1.185$ & $2.146$ & $0.857$ & $0.733$ & $0.862$ & $0.902$ & $0.680$ \\
\quad CatBoost & $1.178$ & $1.951$ & $0.856$ & $0.717$ & $0.866$ & $0.926$ & $0.669$ \\
\quad TabPFN & $1.143$ & $2.009$ & $0.831$ & $0.736$ & $0.854$ & $0.902$ & $0.661$ \\
\quad TabFM & $1.131$ & $1.935$ & $0.847$ & $0.755$ & $0.865$ & $0.914$ & $0.681$ \\
\midrule
\multicolumn{8}{l}{\emph{Molecular graph}} \\
\quad Chemprop & $0.851$ & $2.450$ & $0.676$ & $0.729$ & $0.763$ & $0.875$ & $0.615$ \\
\quad Chemprop, refit on train+valid & $0.839$ & $2.587$ & $0.659$ & $0.723$ & $0.806$ & $0.894$ & $0.639$ \\
\quad Chemprop + rdkit2d & $0.828$ & $2.534$ & $0.727$ & $0.746$ & $0.822$ & $0.904$ & $0.641$ \\
\quad DGT (seed 0) & $0.931$ & $2.666$ & $0.669$ & $0.704$ & $0.847$ & $0.832$ & $0.627$ \\
\quad DGT (published) & $0.925$ & $1.968$ & $0.668$ & $0.737$ & $0.849$ & $0.896$ & $0.650$ \\
\bottomrule
\end{tabular}
\end{table*}

On ESOL and Lipophilicity, the representation effect is larger than the model effect in all
\XsqMain{} main factorial designs, and the difference is significant in each of them
(Table~\ref{tab:squares}). In the TabFM factorial design it exceeds the model effect in
RMSE by \PtFmMaccsInputMinusModelEsol{} [\PtFmMaccsInputMinusModelEsolLo,
\PtFmMaccsInputMinusModelEsolHi] on ESOL and by \PtFmMaccsInputMinusModelLipo{}
[\PtFmMaccsInputMinusModelLipoLo, \PtFmMaccsInputMinusModelLipoHi] on Lipophilicity (full values in
Supplementary S3).
In each of these factorial designs, $R-M$ also exceeds the absolute interaction, so the representation is
the larger effect in either order, although by only \StrictMinMargin{} for \StrictMinModel{} from
\StrictMinStart{} on \StrictMinDataset{}. In every main factorial design it is also the larger on all 10 seeded splits of both datasets. The representation effect is also
larger in the \XsqNewRdkit{} other factorial designs that end at RDKit descriptors, with CatBoost or TabPFN
as the reference model (Supplementary S3). The analysis of
variance gives the same order on ESOL and Lipophilicity, with or without ECFP4 (Supplementary S4).

To check that the result does not depend on MolLogP, MolMR and TPSA, we repeat the main
factorial designs with the remaining 214 RDKit descriptors. The representation effect is still
larger than the model effect in all \XsqRtfN{} main factorial designs, and the difference is
significant in each of them, by \XsqRtfEsolLo--\XsqRtfEsolHi{} RMSE on ESOL and
\XsqRtfLipoLo--\XsqRtfLipoHi{} on Lipophilicity, and by \ExtRtfSolLo--\ExtRtfSolHi{} and
\ExtRtfLipoLo--\ExtRtfLipoHi{} on the external solubility and lipophilicity datasets. The
representation stays the larger effect in either order on both solubility datasets, but on
Lipophilicity only in \XsqRtfLipoStrict{} of the \XsqRtfN{} factorial designs, and in
\ExtRtfLipoStrict{} on the external lipophilicity dataset.

From morgan512 to the RDKit descriptors, the representation effect is larger than the model
effect in all \XsqMgN{} factorial designs between the 4 tabular models, on both ESOL and
Lipophilicity. The difference is significant in all of them, by \XsqMgEsolLo--\XsqMgEsolHi{} RMSE
on ESOL and \XsqMgLipoLo--\XsqMgLipoHi{} on Lipophilicity.
Across the three model pairs tested at both lengths, folding ECFP4 from 2,048 to 512 bits
changed the mean $R-M$ from \XsqMatchedEsolFull{} to \XsqMatchedEsolFolded{} RMSE on ESOL
and from \XsqMatchedLipoFull{} to \XsqMatchedLipoFolded{} RMSE on Lipophilicity
(Tables~\ref{tab:squares}, S4 and S5).

\subsection{Same Result on External Datasets and Low-Similarity Test Molecules}\label{sec:res-external}
On the external datasets, where every model has a higher RMSE (Supplementary S1), the representation effect is again larger than the model effect in all
\XsqMain{} main factorial designs. The difference is significant in each of them, for both solubility and
lipophilicity. For example, on the external solubility dataset, the TabFM factorial design shows
the representation improving the RMSE by \ExtFmInputAvgEsol{} [\ExtFmInputAvgEsolLo,
\ExtFmInputAvgEsolHi] and the model by \ExtFmModelAvgEsol{} [\ExtFmModelAvgEsolLo,
\ExtFmModelAvgEsolHi]. On the external lipophilicity dataset, the same factorial design gives
\ExtFmInputAvgLipo{} [\ExtFmInputAvgLipoLo, \ExtFmInputAvgLipoHi] for the representation and
\ExtFmModelAvgLipo{} [\ExtFmModelAvgLipoLo, \ExtFmModelAvgLipoHi] for the model.
The representation is also the larger effect in either order on the low-similarity halves of the MoleculeNet test sets. It is
larger in \SimInputFar{} of \SimNCells{} cases, the \XsqMain{} main factorial designs on the 3 regression datasets, and in
\SimInputAll{} on the whole test sets (Supplementary S5). Without MolLogP, MolMR and TPSA, the representation is the
larger effect in either order in \SimRtfInputFar{} of the \SimRtfNCells{} cases on the low-similarity halves and in
\SimRtfInputAll{} on the whole test sets.

\subsection{Model Effect and Interaction}\label{sec:res-model}
On solubility and lipophilicity the model effect is smaller than the representation effect, but
TabFM is ahead of the RF on RDKit descriptors on all 7 datasets (sign test,
$p=\SignModel$). On the small
published test sets, this gain is \PtFmMaccsModelRdkitEsol{} RMSE on ESOL, which is not
significant, and a significant \PtFmMaccsModelRdkitLipo{} [\PtFmMaccsModelRdkitLipoLo,
\PtFmMaccsModelRdkitLipoHi] on Lipophilicity. On the external datasets, which hold thousands of
molecules, the gain is significant on both, at \ExtFmModelRdkitEsol{} [\ExtFmModelRdkitEsolLo,
\ExtFmModelRdkitEsolHi] on solubility and \ExtFmModelRdkitLipo{} [\ExtFmModelRdkitLipoLo,
\ExtFmModelRdkitLipoHi] on lipophilicity. Between the two foundation models, TabFM and TabPFN do
not differ significantly on any dataset.

The model effect can also depend on the representation. On Lipophilicity, the model step is
larger on RDKit descriptors than on the fingerprints, and this interaction is significant for TabFM
from MACCS, at \PtFmMaccsInteractionLipo{} [\PtFmMaccsInteractionLipoLo,
\PtFmMaccsInteractionLipoHi], and for TabPFN from ECFP4, at \PtPfnEcfpInteractionLipo{}
[\PtPfnEcfpInteractionLipoLo, \PtPfnEcfpInteractionLipoHi]. The same pattern appears on the
external lipophilicity dataset, where the interaction is positive for TabFM, TabPFN and CatBoost
alike.

\subsection{Result Depends on Dataset and Starting Representation}\label{sec:res-limits}
The FreeSolv answer depends on the starting representation. On FreeSolv, ECFP4 scores far
worse than MACCS and RDKit descriptors (Table~\ref{tab:scores}). Starting from ECFP4, the
representation effect is larger than the model effect in both main factorial designs
(Table~\ref{tab:squares}). For TabPFN the difference is significant, at
\PtPfnEcfpInputMinusModelFreesolv{} [\PtPfnEcfpInputMinusModelFreesolvLo,
\PtPfnEcfpInputMinusModelFreesolvHi] RMSE. From MACCS, the representation step points in opposite ways for the two models: it is
\PtFmMaccsInputRefFreesolv{} for the RF (a loss) but \PtFmMaccsInputCmpFreesolv{} for
TabFM. With only 65 test molecules, the difference between the two effects,
\PtFmMaccsInputMinusModelFreesolv{} [\PtFmMaccsInputMinusModelFreesolvLo,
\PtFmMaccsInputMinusModelFreesolvHi], could be zero. On the
classification datasets only one main factorial design has a significant difference between the two
effects, \ClsSigModel{} from \ClsSigStart{} on \ClsSigDataset{}, where the representation effect is larger by
\ClsSigEst{} AUC. The best representation changes with the dataset (Table~\ref{tab:scores}).

\subsection{Comparison with Published Scores}\label{sec:res-published}
\begin{table}[!htb]
\centering
\footnotesize
\setlength{\tabcolsep}{4pt}
\caption{The baseline of the graph papers with a different representation, on the published split (RMSE, lower is better). RF: our runs. DGT: its published score. Bold: best in the row.}
\label{tab:baseline}
\begin{tabular}{lrrr}
\toprule
Dataset & RF, ECFP4 & RF, RDKit descriptors & DGT \\
\midrule
ESOL & $1.450$ & $\mathbf{0.766}$ & $0.925$ \\
FreeSolv & $3.587$ & $2.256$ & $\mathbf{1.968}$ \\
Lipophilicity & $0.851$ & $0.747$ & $\mathbf{0.668}$ \\
\bottomrule
\end{tabular}
\end{table}

Changing only the representation of the graph papers' RF baseline from ECFP4 to RDKit
descriptors moves it ahead of DGT's published score on ESOL and closes most of the gap on FreeSolv
and Lipophilicity (Table~\ref{tab:baseline}). Over all 7 datasets, the RF on RDKit descriptors is
better than DGT's published score on \NBeatDgtRf{} (Table~\ref{tab:scores}). Chemprop is significantly better than the RF on ECFP4 on all 3
regression datasets, but never significantly better than the RF on RDKit descriptors
(Supplementary S6).

TabFM on RDKit descriptors is ahead of Chemprop on all 7 datasets and significantly on \NGraphSig{}. Adding RDKit descriptors to Chemprop raises its AUC on the \NDescRaiseCls{} classification datasets but
worsens its RMSE on \NDescHurtReg{} of the 3 regression datasets, FreeSolv and Lipophilicity (Table~\ref{tab:scores}), and TabFM is still ahead on all \NAheadDescFm{} datasets and significantly on \NSigDescFm{} (Supplementary S6). Against our DGT runs, TabFM is significantly better on
\NSigDgtFm{} of 7 datasets, while DGT is never significantly better (Supplementary S6). TabFM is
also better than DGT's published score on \NBeatDgtFm{} of 7 datasets, including the
\NFmBeatsPubUnrepro{} where our runs fall short of it. On fingerprints, the tabular foundation
models trail our DGT run in \NFpRegDgtAhead{} of \NFpRegCells{} regression comparisons.

\section{Discussion}

Comparisons against the usual RF on ECFP4 credit the model with gains that come mostly from the
representation, on solubility and lipophilicity. The model effect is
significant on the larger external datasets, but it is smaller than the representation effect.

The two effects also depend on each other. On Lipophilicity, TabFM and TabPFN gain more from RDKit
descriptors than the RF does, so the order of the models can change with the representation. For
example, the RF beats TabPFN on ECFP4, but TabPFN beats the RF on RDKit descriptors
(Table~\ref{tab:scores}). Adding RDKit descriptors to Chemprop does not give it the gain
they give the tabular models on solubility and lipophilicity.

Small test sets leave some cases open. On FreeSolv from MACCS, and on the classification
datasets except in one main factorial design (\ClsSigModel{} from \ClsSigStart{} on
\ClsSigDataset), neither effect is significantly larger, which does not mean the two effects are
equal.

As in earlier work~\cite{deng_systematic_2023,van_tilborg_exposing_2022}, the best representation
depends on what is predicted, a physical property or binding. RDKit descriptors give the best scores
on the 3 physical properties, and ECFP4 on BACE, a binding dataset. The model effect points the same
way as in recent tabular foundation model studies~\cite{chen_tabpfn_2026,matyushin_-context_2026},
although they pooled their results over many datasets instead of testing each one.

The classical
baseline should be given more than one representation, including RDKit descriptors.
When the proposed model
reads a different representation from the baseline, the representation effect, the model effect and their
interaction should be reported. Each dataset also needs a paired, scaffold-aware test, corrected for the number of datasets
tested. The result should then be repeated on other splits and on external molecules. Where no external
data exist, the test molecules least similar to the training set can be used instead.

\section{Limitations}\label{sec:limitations}
\emph{Scope.} Our results come from 7 small datasets, read from the standard DeepChem files and split
by scaffold. They may not hold for larger datasets, similarity-based splits~\cite{fooladi_evaluating_2025}
or cleaned data. The external datasets are predicted by the models trained on the published split, and both
external datasets mix measurement protocols. Differences between these protocols may contribute to the
higher external RMSE. We cannot separate their contribution from that of differences between the training
and external molecules. The external solubility dataset checks a new source
rather than unfamiliar molecules, because its molecules are about as similar to ESOL's training
molecules as ESOL's own test set (Supplementary S1).

\emph{Factorial designs.} A more fully tuned reference model or a larger descriptor set could change the
effects. With the 2,048 bits of ECFP4, TabFM's GPU memory use exceeds what the larger datasets
allow, so it does not run on all 7 datasets. Its factorial design therefore starts
from MACCS and not from the usual ECFP4 baseline. TabPFN does read ECFP4, but on 4 datasets the fingerprint has more features than TabPFN's
stated limit.

\emph{Graph models.} Chemprop is tuned over only 4 settings, so a more fully tuned Chemprop may score better. Our
tests against DGT use the seed-0 run on the published split, so they do not cover the variation between DGT's training
runs. Seeds 1--3 are in Supplementary S6.

\section{Conclusion}
Most of the gain that tabular models show over the usual RF baseline comes from the representation.
This holds on small solubility and lipophilicity datasets when the RF reads ECFP4 or MACCS. On
FreeSolv from MACCS and on most classification designs, neither effect is significantly larger. The RF reaches much of the representation's gain once it
reads RDKit descriptors instead of a fingerprint. The model adds a smaller gain, and the best
representation changed with the model and the dataset. For small physical-property datasets under
scaffold splits, a next step is to treat the choice of molecular representation like a hyperparameter
and select it for each model and dataset using validation performance.

\end{document}


\title{Supplementary Material: Representation and Model Contributions on Small Molecular Benchmarks}
\author{\IEEEauthorblockN{Nora A. M. A. Talkhan}
\IEEEauthorblockA{\textit{Faculty of Computing and}\\
\textit{Artificial Intelligence}\\
\textit{Cairo University}, Giza, Egypt\\
n.abdelhameed@fci-cu.edu.eg}
\and[\hspace{20pt}]
\IEEEauthorblockN{Ibrahim Abdelbaky}
\IEEEauthorblockA{\textit{Faculty of Computers and Information Technology}\\
\textit{Future University}, Cairo, Egypt\\
\textit{Faculty of Computers and Artificial Intelligence}\\
\textit{Benha University}, Banha, Egypt\\
Ibrahim.Abdelbaky@fue.edu.eg}
\and[\hspace{20pt}]
\IEEEauthorblockN{Fatma A. Omara}
\IEEEauthorblockA{\textit{Faculty of Computing and}\\
\textit{Artificial Intelligence}\\
\textit{Cairo University}, Giza, Egypt\\
f.omara@fci-cu.edu.eg}}
\maketitle
\raggedbottom

\section*{S1. Datasets and Statistical Tests}
The size of each dataset and of the published split is in Table~\ref{tab:data}.

\emph{External datasets.} The AqSolDBc authors describe the solubility values as thermodynamic
aqueous solubilities at pH $7 \pm 1$ and approximately 300~K. For the RTlogD external test set,
the authors report logD measurements at pH 7.2--7.6 obtained by the shake-flask method,
chromatography or potentiometric titration.
From each external dataset we remove every molecule that also occurs
anywhere in the MoleculeNet dataset, by InChIKey or canonical tautomer. We also remove molecules with
elements absent from the training data, and repeated molecules whose labels differ by more than 0.5
log units.
For every tabular model and representation we test, RMSE is higher on the external dataset
than on the corresponding MoleculeNet test set (Table~\ref{tab:extabs}).

For each test molecule, we find the most similar molecule in the training or validation set
of the published split, using ECFP4 Tanimoto similarity. For solubility, the median similarity is
\ExtSimEsol{} on the external set and \ExtSimEsolMN{} on the ESOL test set. For lipophilicity,
the medians are \ExtSimLipo{} on the external set and \ExtSimLipoMN{} on the MoleculeNet test set.

\emph{Statistical tests.} The two-sided bootstrap p-value is $\min(1,2(k+1)/(N+1))$, where
$N$ is the number of draws. We count the draws at or below zero and the draws at or above zero.
The smaller count is $k$. The sign test across datasets uses only the direction of each
difference. With 7 datasets, its smallest possible p-value is 0.016.

\begin{table}[h]
\centering
\caption{Datasets and the published split.}
\label{tab:data}
\begin{tabular}{lrrrrr}
\toprule
Dataset & Molecules & Tasks & Train & Valid & Test \\
\midrule
ESOL          & 1,128 & 1  &   902 & 113 & 113 \\
FreeSolv      &   642 & 1  &   513 &  64 &  65 \\
Lipophilicity & 4,200 & 1  & 3,360 & 420 & 420 \\
BBBP          & 2,039 & 1  & 1,631 & 204 & 204 \\
BACE          & 1,513 & 1  & 1,210 & 151 & 152 \\
ClinTox       & 1,480 & 2  & 1,184 & 148 & 148 \\
SIDER         & 1,427 & 27 & 1,141 & 143 & 143 \\
\bottomrule
\end{tabular}
\end{table}

\section*{S2. Molecular Identity Check Across Splits}
We check whether test molecules also appear in the training or validation data.
We match molecules by the InChIKey of the parent structure or by their canonical tautomer.
We find matches for \IdentBbbp{} of \IdentTestBbbp{} test molecules in BBBP,
\IdentBace{} of \IdentTestBace{} in BACE and \IdentSider{} of \IdentTestSider{} in SIDER.
In \IdentLabelDiffers{} cases, the labels differ. For example, BBBP lists theophylline twice,
once with a positive label and once with a negative label.
We remove the matching test entries and score the saved predictions again.
In the main comparisons, the largest change is \LeakMaxChange{} and no difference changes sign.

\section*{S3. Results for All Factorial Designs}
Tables~\ref{tab:parts} and~\ref{tab:othersquares} give the \XsqMain{} main designs and
\XsqNew{} additional designs. The additional designs use CatBoost or TabPFN as the reference
model or compare ECFP4 with MACCS. They reuse runs from the main designs, so the results
are not independent. Each comparison starts with the earlier model (RF, CatBoost, TabPFN, TabFM)
and the earlier representation (ECFP4, MACCS, RDKit descriptors).
Reversing both comparisons changes the signs of the representation and model effects but preserves
their magnitudes.

We repeated the main designs after removing MolLogP, MolMR and TPSA. We also compared the
512-bit Morgan fingerprint with the full RDKit descriptor set for all \XsqMgN{} model pairs
(Table~\ref{tab:restsquares}). Switching from a fingerprint to
RDKit descriptors changed each model's score as shown in Table~\ref{tab:side}.

In the \XsqNewRdkit{} other factorial designs that end at RDKit descriptors (Table~\ref{tab:othersquares}), the representation effect is
\XsqInEsolLo--\XsqInEsolHi{} RMSE on ESOL and \XsqInLipoLo--\XsqInLipoHi{} on Lipophilicity, and no
model effect is larger than \XsqMoMax{} in size. The representation effect is larger on all
\XsqSeedMin{} seeded splits.

On FreeSolv, RDKit descriptors instead of MACCS make the RF worse (\XsqFsRf{} on the
published split), change CatBoost in no consistent direction (\XsqFsCat{} on the published split,
\XsqFsCatSeed{} on average over the seeded splits) and improve TabPFN and TabFM
(\XsqFsGainLo--\XsqFsGainHi, Table~\ref{tab:side}). The seeded splits give the same signs
for the RF, TabPFN and TabFM, but none of these differences is significant.

Among the \XsqNew{} other factorial designs in Table~\ref{tab:othersquares}, a single one has a
significant difference on the classification datasets, on \XsqClsSigSets.
It runs from CatBoost on ECFP4 to TabPFN on RDKit descriptors. The matching main factorial design with the RF as
reference model has similar differences that are not significant (Table II of the main paper), so we
treat this result as weak.

MACCS is better than ECFP4 on ESOL and FreeSolv for all 3 models that read ECFP4, and significantly
so on FreeSolv (\XsqEmFsLo--\XsqEmFsHi{} RMSE).

Across the \XsqRepChangeN{} designs in Tables~\ref{tab:parts} and~\ref{tab:othersquares}, the
representation effect is significantly larger in \XsqCsRegIn{} of \XsqCsRegN{} regression
comparisons and \XsqCsClsIn{} of \XsqCsClsN{} classification comparisons. None of these
comparisons shows a significantly larger model effect.

On both external datasets, the average representation effect is larger than the model effect
in all \XsqNew{} additional designs. The difference is \XextSigText.

From morgan512 to the 214 RDKit descriptors, $R-M$ is positive on ESOL and Lipophilicity
in all \XsqMgRtfN{} factorial designs, in all three evaluations.

\section*{S4. Full Factorial Analysis of Variance}
The analysis uses the split as a block. Each share is the part of the variation within splits that a
term explains, with a 95\% interval from \VarNBoot{} bootstrap draws of the 10 splits.

\begin{figure*}[t]
\centering
\includegraphics[width=0.72\textwidth]{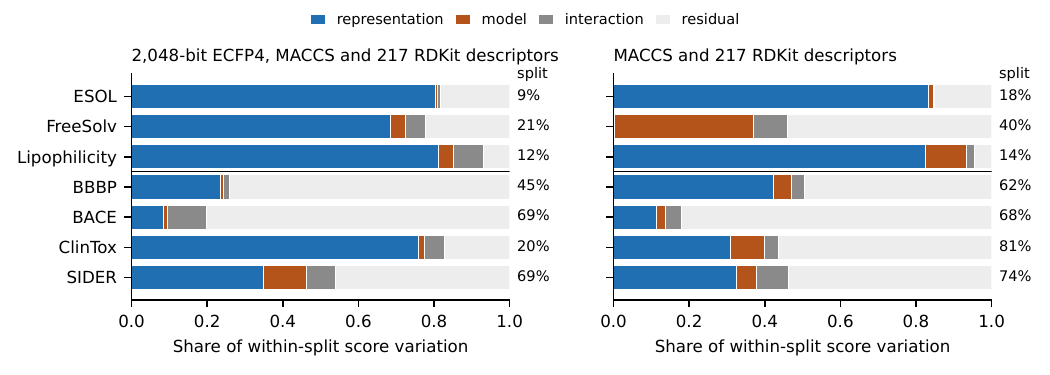}
\caption{Both panels compare RF, CatBoost and TabPFN across ten scaffold splits.
The bars show how much of the score variation within splits comes from the representation,
the model and their interaction. The rest is residual variation. The percentages beside
the bars give the split's share of the total variation.}
\label{fig:variance}
\end{figure*}

With ECFP4, MACCS and RDKit descriptors, the representation share is significantly larger than
the model share on every dataset except \VarANotSigSets{} (Fig.~\ref{fig:variance}).
With only MACCS and the full set of RDKit descriptors, it stays significantly larger on \VarDSigSets{}.
On \VarLostSets{}, removing ECFP4 leaves no significant difference between the representation and model shares.

On FreeSolv, the choice of model explains \VarDFsMo{} of the variation within splits.
Representation alone explains less than 1\%. The RF does better with MACCS,
while TabPFN does better with RDKit descriptors. These differences largely cancel when averaged
across models. More than half of the variation within splits remains unexplained.
When we add the 512-bit Morgan fingerprint as a third representation, the choice of representation
explains more variation than the choice of model on FreeSolv.

\section*{S5. High- and Low-Similarity Test Molecules}
We split each test set into two halves by similarity to the training and validation molecules.
Changing the representation helps more than changing the model in either order in
\SimInputNear{} of \SimNCells{} comparisons in the more similar half and \SimInputFar{} of
\SimNCells{} in the less similar half (Table~\ref{tab:similarity}). Only Lipophilicity differs
between halves in how many designs meet this criterion. The gap between the representation
and model effects differs significantly between halves in only \SimNResFarMinusNear{} of
\SimNFarMinusNear{} comparisons.

\section*{S6. Model Comparisons and DGT Reproduction}
Tables~\ref{tab:comparisons} and~\ref{tab:chemprop} give the comparisons between models.

\emph{DGT.} Our seed-0 runs reach DGT's published scores on ESOL, Lipophilicity and BACE. On FreeSolv, BBBP, ClinTox
and SIDER, all our runs, including seeds 1--3 and the settings of DGT's Zenodo release, are worse than the published values
(FreeSolv \DgtRunsLoFreesolv--\DgtRunsHiFreesolv{} RMSE, BBBP \DgtRunsLoBbbp--\DgtRunsHiBbbp, ClinTox
\DgtRunsLoClintox--\DgtRunsHiClintox{} and SIDER \DgtRunsLoSider--\DgtRunsHiSider{} AUC). We do not
know the cause of the gap. Table~\ref{tab:dgt} shows the DGT run with seed 0.
TabFM on RDKit descriptors scores significantly better than DGT for all \DgtSeedN{} seeds on
ESOL, FreeSolv, ClinTox and SIDER. On BBBP, it does so for seeds \DgtBbbpWinningSeeds{}.
Neither model has a significant advantage on Lipophilicity or BACE.
DGT has no significant wins against TabFM on RDKit descriptors.

\emph{Chemprop variants and TabFM+.} On the seeded splits, Chemprop beats Chemprop + rdkit2d on \WDescGainEsol{} of 10 splits on ESOL, \WDescGainFreesolv{} on
FreeSolv, \WDescGainLipo{} on Lipophilicity and \WDescGainBace{} on BACE. Adding the validation molecules helps Chemprop on \NRefitWinsSeed{} of the \NRefitSplitsSeed{} seeded splits, but the difference is not significant on any dataset on the published split. TabFM+ \PlusFmVerdict.

\section*{S7. Run Time and Memory}
We time each model on ESOL and Lipophilicity using one process per model and dataset
on a machine with \TimeCpus{} CPU cores and a \TimeGpuMem{} GB A100 GPU.
Table~\ref{tab:timing} reports fit time, prediction time and peak memory.
All models take longer on Lipophilicity than on ESOL when fitting and prediction are timed together.

Across the two datasets, TabPFN on the GPU takes \TimePfnRfLo{} to \TimePfnRfHi{} times as long
as the RF on the CPU for fitting and prediction combined. TabFM takes \TimeFmRfLo{} to
\TimeFmRfHi{} times as long as the RF and spends most of that time on prediction.
Peak RAM use is higher for TabPFN and TabFM than for the RF on both datasets.
TabFM uses \TimeFmRamMeasured{} GB of RAM on both datasets. Its peak GPU memory rises from
\TimeFmGpuEsol{} GB on ESOL to \TimeFmGpuLipo{} GB on Lipophilicity.

Both datasets have one task. On datasets with several tasks, the tabular models run once
for each task. Chemprop fits all tasks together.

\begin{table*}[t]
\centering
\footnotesize
\setlength{\tabcolsep}{4pt}
\caption{The 5 main factorial designs, published split. Each design runs from the RF to the named model and from the named representation to rdkit2d. Positive values are improvements. Brackets give the 95\% interval, bold values are significant, and S counts the seeded splits (of 10) where the representation effect is larger.}
\label{tab:parts}
\resizebox{\textwidth}{!}{\begin{tabular}{lrrrrr}
\toprule
Dataset & Repr.\ effect & Model effect & Repr.\ $-$ model & Interaction & S \\
\midrule
\multicolumn{6}{l}{\emph{RF $\to$ TabFM, MACCS $\to$ rdkit2d}} \\
\quad ESOL (RMSE) & $\mathbf{0.417}$ [$0.264$, $0.572$] & $0.053$ [$-0.020$, $0.124$] & $\mathbf{0.364}$ [$0.210$, $0.515$] & $-0.004$ [$-0.119$, $0.108$] & 10 \\
\quad FreeSolv (RMSE) & $0.043$ [$-0.430$, $0.463$] & $0.364$ [$0.060$, $0.605$] & $-0.321$ [$-0.815$, $0.108$] & $0.307$ [$-0.435$, $1.052$] & 4 \\
\quad Lipophilicity (RMSE) & $\mathbf{0.155}$ [$0.103$, $0.209$] & $\mathbf{0.055}$ [$0.022$, $0.087$] & $\mathbf{0.101}$ [$0.043$, $0.158$] & $\mathbf{0.090}$ [$0.038$, $0.141$] & 10 \\
\quad BBBP (AUC) & $0.009$ [$-0.022$, $0.042$] & $0.015$ [$-0.007$, $0.038$] & $-0.006$ [$-0.047$, $0.036$] & $-0.015$ [$-0.049$, $0.019$] & 6 \\
\quad BACE (AUC) & $-0.031$ [$-0.073$, $0.008$] & $0.016$ [$-0.012$, $0.044$] & $-0.047$ [$-0.098$, $0.002$] & $0.026$ [$-0.014$, $0.067$] & 6 \\
\quad ClinTox (AUC) & $0.011$ [$-0.023$, $0.044$] & $0.035$ [$-0.005$, $0.080$] & $-0.024$ [$-0.067$, $0.017$] & $0.044$ [$-0.003$, $0.101$] & 0 \\
\quad SIDER (AUC) & $0.011$ [$-0.011$, $0.035$] & $0.004$ [$-0.013$, $0.021$] & $0.008$ [$-0.021$, $0.035$] & $0.005$ [$-0.025$, $0.030$] & 7 \\
\midrule
\multicolumn{6}{l}{\emph{RF $\to$ TabPFN, MACCS $\to$ rdkit2d}} \\
\quad ESOL (RMSE) & $\mathbf{0.414}$ [$0.267$, $0.559$] & $0.038$ [$-0.030$, $0.104$] & $\mathbf{0.377}$ [$0.235$, $0.515$] & $-0.009$ [$-0.102$, $0.082$] & 10 \\
\quad FreeSolv (RMSE) & $0.066$ [$-0.521$, $0.535$] & $0.313$ [$0.078$, $0.512$] & $-0.247$ [$-0.939$, $0.297$] & $0.353$ [$-0.083$, $0.756$] & 3 \\
\quad Lipophilicity (RMSE) & $\mathbf{0.148}$ [$0.093$, $0.204$] & $\mathbf{0.063}$ [$0.034$, $0.092$] & $\mathbf{0.084}$ [$0.029$, $0.141$] & $0.074$ [$0.020$, $0.127$] & 10 \\
\quad BBBP (AUC) & $0.023$ [$-0.008$, $0.057$] & $0.009$ [$-0.011$, $0.030$] & $0.013$ [$-0.026$, $0.054$] & $0.013$ [$-0.019$, $0.047$] & 7 \\
\quad BACE (AUC) & $-0.025$ [$-0.069$, $0.017$] & $0.011$ [$-0.015$, $0.038$] & $-0.036$ [$-0.090$, $0.018$] & $0.039$ [$-0.002$, $0.081$] & 5 \\
\quad ClinTox (AUC) & $0.012$ [$-0.018$, $0.042$] & $0.024$ [$-0.009$, $0.058$] & $-0.012$ [$-0.053$, $0.031$] & $0.047$ [$0.005$, $0.092$] & 0 \\
\quad SIDER (AUC) & $0.019$ [$-0.005$, $0.042$] & $-0.009$ [$-0.026$, $0.008$] & $0.028$ [$0.000$, $0.056$] & $0.019$ [$-0.004$, $0.041$] & 10 \\
\midrule
\multicolumn{6}{l}{\emph{RF $\to$ CatBoost, MACCS $\to$ rdkit2d}} \\
\quad ESOL (RMSE) & $\mathbf{0.429}$ [$0.275$, $0.585$] & $0.017$ [$-0.027$, $0.064$] & $\mathbf{0.412}$ [$0.254$, $0.569$] & $0.020$ [$-0.052$, $0.102$] & 10 \\
\quad FreeSolv (RMSE) & $-0.096$ [$-0.778$, $0.453$] & $0.209$ [$0.045$, $0.356$] & $-0.305$ [$-0.970$, $0.266$] & $0.029$ [$-0.221$, $0.305$] & 2 \\
\quad Lipophilicity (RMSE) & $\mathbf{0.134}$ [$0.080$, $0.190$] & $0.025$ [$-0.004$, $0.054$] & $\mathbf{0.110}$ [$0.046$, $0.176$] & $0.048$ [$-0.007$, $0.099$] & 10 \\
\quad BBBP (AUC) & $0.015$ [$-0.021$, $0.051$] & $-0.018$ [$-0.035$, $-0.001$] & $0.032$ [$-0.007$, $0.074$] & $-0.003$ [$-0.034$, $0.028$] & 8 \\
\quad BACE (AUC) & $-0.046$ [$-0.094$, $-0.001$] & $0.003$ [$-0.025$, $0.032$] & $-0.049$ [$-0.104$, $0.004$] & $-0.004$ [$-0.050$, $0.041$] & 5 \\
\quad ClinTox (AUC) & $-0.012$ [$-0.059$, $0.028$] & $0.025$ [$-0.002$, $0.053$] & $-0.036$ [$-0.093$, $0.008$] & $0.000$ [$-0.046$, $0.049$] & 2 \\
\quad SIDER (AUC) & $0.013$ [$-0.011$, $0.037$] & $-0.007$ [$-0.022$, $0.006$] & $0.020$ [$-0.009$, $0.049$] & $0.007$ [$-0.018$, $0.032$] & 8 \\
\midrule
\multicolumn{6}{l}{\emph{RF $\to$ TabPFN, ECFP4 $\to$ rdkit2d}} \\
\quad ESOL (RMSE) & $\mathbf{0.662}$ [$0.486$, $0.838$] & $0.055$ [$-0.007$, $0.118$] & $\mathbf{0.607}$ [$0.434$, $0.779$] & $-0.044$ [$-0.140$, $0.052$] & 10 \\
\quad FreeSolv (RMSE) & $\mathbf{1.312}$ [$0.787$, $1.839$] & $\mathbf{0.508}$ [$0.207$, $0.747$] & $\mathbf{0.804}$ [$0.295$, $1.309$] & $-0.038$ [$-0.678$, $0.596$] & 10 \\
\quad Lipophilicity (RMSE) & $\mathbf{0.181}$ [$0.130$, $0.231$] & $0.024$ [$-0.005$, $0.052$] & $\mathbf{0.157}$ [$0.103$, $0.209$] & $\mathbf{0.153}$ [$0.102$, $0.202$] & 10 \\
\quad BBBP (AUC) & $0.050$ [$0.004$, $0.097$] & $0.011$ [$-0.011$, $0.034$] & $0.039$ [$-0.015$, $0.095$] & $0.010$ [$-0.040$, $0.060$] & 7 \\
\quad BACE (AUC) & $-0.046$ [$-0.094$, $-0.003$] & $0.010$ [$-0.011$, $0.031$] & $-0.056$ [$-0.112$, $-0.003$] & $0.042$ [$0.005$, $0.081$] & 6 \\
\quad ClinTox (AUC) & $\mathbf{0.151}$ [$0.018$, $0.359$] & $-0.011$ [$-0.086$, $0.046$] & $0.162$ [$0.006$, $0.364$] & $0.118$ [$0.002$, $0.254$] & 10 \\
\quad SIDER (AUC) & $0.030$ [$0.000$, $0.061$] & $-0.009$ [$-0.024$, $0.007$] & $0.039$ [$0.004$, $0.074$] & $0.018$ [$-0.016$, $0.051$] & 9 \\
\midrule
\multicolumn{6}{l}{\emph{RF $\to$ CatBoost, ECFP4 $\to$ rdkit2d}} \\
\quad ESOL (RMSE) & $\mathbf{0.676}$ [$0.511$, $0.846$] & $0.035$ [$-0.021$, $0.091$] & $\mathbf{0.641}$ [$0.471$, $0.816$] & $-0.015$ [$-0.147$, $0.117$] & 10 \\
\quad FreeSolv (RMSE) & $\mathbf{1.237}$ [$0.509$, $1.944$] & $\mathbf{0.318}$ [$0.153$, $0.469$] & $0.919$ [$0.154$, $1.637$] & $-0.188$ [$-0.443$, $0.074$] & 9 \\
\quad Lipophilicity (RMSE) & $\mathbf{0.116}$ [$0.067$, $0.166$] & $0.037$ [$0.007$, $0.064$] & $\mathbf{0.080}$ [$0.023$, $0.136$] & $0.024$ [$-0.024$, $0.069$] & 10 \\
\quad BBBP (AUC) & $0.042$ [$-0.009$, $0.094$] & $-0.016$ [$-0.038$, $0.005$] & $0.058$ [$0.001$, $0.117$] & $-0.006$ [$-0.048$, $0.035$] & 7 \\
\quad BACE (AUC) & $-0.034$ [$-0.087$, $0.018$] & $-0.033$ [$-0.068$, $0.000$] & $-0.001$ [$-0.073$, $0.073$] & $0.067$ [$0.009$, $0.126$] & 6 \\
\quad ClinTox (AUC) & $0.109$ [$-0.023$, $0.308$] & $0.007$ [$-0.052$, $0.066$] & $0.102$ [$-0.034$, $0.267$] & $0.035$ [$-0.065$, $0.144$] & 10 \\
\quad SIDER (AUC) & $0.035$ [$0.004$, $0.067$] & $-0.018$ [$-0.033$, $-0.004$] & $\mathbf{0.054}$ [$0.016$, $0.092$] & $0.029$ [$-0.007$, $0.062$] & 10 \\
\bottomrule
\end{tabular}}
\end{table*}

\begin{table*}[t]
\centering
\footnotesize
\caption{RMSE on the MoleculeNet test sets and the external datasets.}
\label{tab:extabs}
\begin{tabular}{llrrrr}
\toprule
 & & \multicolumn{2}{c}{Solubility} & \multicolumn{2}{c}{Lipophilicity} \\
\cmidrule(lr){3-4} \cmidrule(lr){5-6}
Model & Representation & ESOL test & External & Lipophilicity test & External \\
\midrule
RF & ECFP4 & $1.450$ & $1.744$ & $0.851$ & $1.418$ \\
RF & MACCS & $1.185$ & $1.645$ & $0.857$ & $1.349$ \\
RF & morgan512 & $1.491$ & $1.784$ & $0.927$ & $1.404$ \\
RF & RDKit descriptors & $0.766$ & $1.323$ & $0.747$ & $1.177$ \\
RF & 214 RDKit descriptors & $0.860$ & $1.460$ & $0.744$ & $1.255$ \\
CatBoost & ECFP4 & $1.408$ & $1.711$ & $0.826$ & $1.356$ \\
CatBoost & MACCS & $1.178$ & $1.588$ & $0.856$ & $1.335$ \\
CatBoost & morgan512 & $1.470$ & $1.758$ & $0.867$ & $1.352$ \\
CatBoost & RDKit descriptors & $0.739$ & $1.260$ & $0.698$ & $1.102$ \\
CatBoost & 214 RDKit descriptors & $0.825$ & $1.331$ & $0.706$ & $1.137$ \\
TabPFN & ECFP4 & $1.373$ & $1.745$ & $0.903$ & $1.382$ \\
TabPFN & MACCS & $1.143$ & $1.591$ & $0.831$ & $1.321$ \\
TabPFN & morgan512 & $1.318$ & $1.732$ & $0.835$ & $1.340$ \\
TabPFN & RDKit descriptors & $0.733$ & $1.224$ & $0.646$ & $1.081$ \\
TabPFN & 214 RDKit descriptors & $0.752$ & $1.283$ & $0.630$ & $1.082$ \\
TabFM & MACCS & $1.131$ & $1.595$ & $0.847$ & $1.305$ \\
TabFM & morgan512 & $1.398$ & $1.777$ & $0.894$ & $1.341$ \\
TabFM & RDKit descriptors & $0.715$ & $1.244$ & $0.647$ & $1.086$ \\
TabFM & 214 RDKit descriptors & $0.730$ & $1.325$ & $0.653$ & $1.108$ \\
\bottomrule
\end{tabular}
\end{table*}

\begin{table*}[t]
\centering
\footnotesize
\setlength{\tabcolsep}{4pt}
\caption{Representation effect $-$ model effect in the 7 other factorial designs, published split. Each column gives the reference and compared model, then the starting and final representation. Positive values mean the representation helped more, and bold values are significant.}
\label{tab:othersquares}
\begin{tabular}{lrrrrrrr}
\toprule
Models & \begin{tabular}[b]{@{}r@{}}CatBoost\\$\to$TabPFN\end{tabular} & \begin{tabular}[b]{@{}r@{}}CatBoost\\$\to$TabFM\end{tabular} & \begin{tabular}[b]{@{}r@{}}TabPFN\\$\to$TabFM\end{tabular} & \begin{tabular}[b]{@{}r@{}}CatBoost\\$\to$TabPFN\end{tabular} & \begin{tabular}[b]{@{}r@{}}RF\\$\to$TabPFN\end{tabular} & \begin{tabular}[b]{@{}r@{}}RF\\$\to$CatBoost\end{tabular} & \begin{tabular}[b]{@{}r@{}}CatBoost\\$\to$TabPFN\end{tabular} \\
\addlinespace
Representations & \begin{tabular}[b]{@{}r@{}}MACCS\\$\to$rdkit2d\end{tabular} & \begin{tabular}[b]{@{}r@{}}MACCS\\$\to$rdkit2d\end{tabular} & \begin{tabular}[b]{@{}r@{}}MACCS\\$\to$rdkit2d\end{tabular} & \begin{tabular}[b]{@{}r@{}}ECFP4\\$\to$rdkit2d\end{tabular} & \begin{tabular}[b]{@{}r@{}}ECFP4\\$\to$MACCS\end{tabular} & \begin{tabular}[b]{@{}r@{}}ECFP4\\$\to$MACCS\end{tabular} & \begin{tabular}[b]{@{}r@{}}ECFP4\\$\to$MACCS\end{tabular} \\
\midrule
ESOL (RMSE) & $\mathbf{0.404}$ & $\mathbf{0.392}$ & $\mathbf{0.398}$ & $\mathbf{0.634}$ & $0.188$ & $0.222$ & $0.195$ \\
FreeSolv (RMSE) & $-0.023$ & $-0.097$ & $0.168$ & $\mathbf{1.028}$ & $\mathbf{0.914}$ & $\mathbf{1.029}$ & $\mathbf{1.109}$ \\
Lipophilicity (RMSE) & $\mathbf{0.133}$ & $\mathbf{0.149}$ & $\mathbf{0.201}$ & $\mathbf{0.205}$ & $0.046$ & $-0.031$ & $0.047$ \\
BBBP (AUC) & $-0.006$ & $-0.025$ & $0.011$ & $0.020$ & $0.023$ & $0.042$ & $0.007$ \\
BACE (AUC) & $-0.035$ & $-0.047$ & $-0.017$ & $-0.055$ & $-0.012$ & $0.043$ & $-0.007$ \\
ClinTox (AUC) & $0.012$ & $0.000$ & $0.023$ & $\mathbf{0.187}$ & $0.174$ & $0.114$ & $0.199$ \\
SIDER (AUC) & $0.024$ & $0.004$ & $0.008$ & $0.035$ & $0.029$ & $0.045$ & $0.018$ \\
\bottomrule
\end{tabular}
\end{table*}

\begin{table*}[t]
\centering
\footnotesize
\caption{Representation effect $-$ model effect in the descriptor ablation and 512-bit fingerprint designs, published split. Each row gives the reference and compared model, under its starting and final representation. rdkit2d\_214 is rdkit2d without MolLogP, MolMR and TPSA. Positive values mean the representation helped more, and bold values are significant.}
\label{tab:restsquares}
\begin{tabular}{lrrrrrrr}
\toprule
 & \multicolumn{3}{c}{Regression: RMSE $\downarrow$} & \multicolumn{4}{c}{Classification: ROC-AUC $\uparrow$} \\ \cmidrule(lr){2-4} \cmidrule(lr){5-8}
Models & ESOL & FreeSolv & Lipo & BBBP & BACE & ClinTox & SIDER \\
\midrule
\multicolumn{8}{l}{\emph{ECFP4 $\to$ rdkit2d\_214}} \\
RF $\to$ CatBoost & $\mathbf{0.547}$ & $\mathbf{0.924}$ & $\mathbf{0.082}$ & $0.059$ & $-0.009$ & $0.104$ & $\mathbf{0.054}$ \\
RF $\to$ TabPFN & $\mathbf{0.513}$ & $\mathbf{0.809}$ & $\mathbf{0.159}$ & $0.040$ & $-0.064$ & $0.164$ & $0.039$ \\
\multicolumn{8}{l}{\emph{MACCS $\to$ rdkit2d\_214}} \\
RF $\to$ CatBoost & $\mathbf{0.318}$ & $-0.299$ & $\mathbf{0.112}$ & $0.034$ & $-0.057$ & $-0.035$ & $0.020$ \\
RF $\to$ TabPFN & $\mathbf{0.283}$ & $-0.241$ & $\mathbf{0.086}$ & $0.015$ & $-0.044$ & $-0.011$ & $0.029$ \\
RF $\to$ TabFM & $\mathbf{0.270}$ & $-0.315$ & $\mathbf{0.103}$ & $-0.004$ & $-0.055$ & $-0.023$ & $0.008$ \\
\multicolumn{8}{l}{\emph{morgan512 $\to$ rdkit2d}} \\
RF $\to$ CatBoost & $\mathbf{0.704}$ & $0.707$ & $\mathbf{0.121}$ & $-0.001$ & $-0.032$ & $0.121$ & $0.036$ \\
RF $\to$ TabPFN & $\mathbf{0.552}$ & $\mathbf{0.879}$ & $\mathbf{0.089}$ & $0.019$ & $-0.056$ & $0.117$ & $0.026$ \\
RF $\to$ TabFM & $\mathbf{0.631}$ & $\mathbf{0.641}$ & $\mathbf{0.147}$ & $0.026$ & $-0.035$ & $0.135$ & $0.042$ \\
CatBoost $\to$ TabPFN & $\mathbf{0.579}$ & $\mathbf{1.104}$ & $\mathbf{0.137}$ & $0.000$ & $-0.055$ & $0.141$ & $0.022$ \\
CatBoost $\to$ TabFM & $\mathbf{0.659}$ & $\mathbf{0.865}$ & $\mathbf{0.196}$ & $0.007$ & $-0.034$ & $0.159$ & $0.038$ \\
TabPFN $\to$ TabFM & $\mathbf{0.665}$ & $\mathbf{1.130}$ & $\mathbf{0.247}$ & $0.043$ & $-0.004$ & $0.182$ & $0.042$ \\
\bottomrule
\end{tabular}
\end{table*}

\begin{table*}[t]
\centering
\footnotesize
\caption{Each model on rdkit2d against the same model on a fingerprint, published split. A positive difference means rdkit2d is better. Bold values are significant, and brackets count the seeded splits (of 10) where rdkit2d is better.}
\label{tab:side}
\setlength{\tabcolsep}{4pt}
\begin{tabular}{lrrrrrrr}
\toprule
 & \multicolumn{2}{c}{RF} & \multicolumn{2}{c}{TabPFN} & \multicolumn{1}{c}{TabFM} & \multicolumn{2}{c}{CatBoost} \\
\cmidrule(lr){2-3} \cmidrule(lr){4-5} \cmidrule(lr){6-6} \cmidrule(lr){7-8}
Dataset & ECFP4 & MACCS & ECFP4 & MACCS & MACCS & ECFP4 & MACCS \\
\midrule
ESOL (RMSE) & $\mathbf{0.684}$ (10) & $\mathbf{0.419}$ (10) & $\mathbf{0.640}$ (10) & $\mathbf{0.410}$ (10) & $\mathbf{0.415}$ (10) & $\mathbf{0.669}$ (10) & $\mathbf{0.439}$ (10) \\
FreeSolv (RMSE) & $\mathbf{1.331}$ (10) & $-0.110$ (4) & $\mathbf{1.293}$ (10) & $0.243$ (7) & $0.196$ (7) & $\mathbf{1.143}$ (10) & $-0.081$ (7) \\
Lipophilicity (RMSE) & $\mathbf{0.104}$ (10) & $\mathbf{0.110}$ (10) & $\mathbf{0.257}$ (10) & $\mathbf{0.185}$ (10) & $\mathbf{0.200}$ (10) & $\mathbf{0.128}$ (10) & $\mathbf{0.158}$ (10) \\
BBBP (AUC) & $0.045$ (8) & $0.016$ (9) & $0.055$ (8) & $0.030$ (10) & $0.001$ (9) & $0.039$ (8) & $0.013$ (7) \\
BACE (AUC) & $\mathbf{-0.067}$ (2) & $-0.044$ (6) & $-0.026$ (9) & $-0.005$ (8) & $-0.018$ (8) & $0.000$ (6) & $-0.048$ (4) \\
ClinTox (AUC) & $0.092$ (9) & $-0.012$ (1) & $\mathbf{0.210}$ (10) & $0.035$ (2) & $0.033$ (7) & $0.127$ (10) & $-0.012$ (1) \\
SIDER (AUC) & $0.021$ (8) & $0.009$ (7) & $0.039$ (9) & $0.028$ (10) & $0.014$ (7) & $\mathbf{0.050}$ (9) & $0.016$ (6) \\
\bottomrule
\end{tabular}
\end{table*}

\begin{table*}[t]
\centering
\footnotesize
\setlength{\tabcolsep}{5pt}
\caption{High- and low-similarity halves of each test set, published split. The steps are those of the TabFM factorial design, in RMSE, and positive values are improvements. Similarity is the median highest ECFP4 Tanimoto similarity to the training and validation molecules. Bold values are significant. The last 2 columns count the main factorial designs in which each representation step is larger than each model step.}
\label{tab:similarity}
\begin{tabular}{llrrrrrrrr}
\toprule
 & & & & \multicolumn{2}{c}{Representation step} & \multicolumn{2}{c}{Model step} & \multicolumn{2}{c}{Designs (of 5)} \\
\cmidrule(lr){5-6} \cmidrule(lr){7-8} \cmidrule(lr){9-10}
Dataset & Half & Molecules & Similarity & within RF & within TabFM & on MACCS & on rdkit2d & Published & Seeded (mean) \\
\midrule
ESOL & high & 57 & $0.50$ & $\mathbf{0.632}$ & $\mathbf{0.603}$ & $0.091$ & $0.061$ & 5 & $5.0$ \\
 & low & 56 & $0.30$ & $0.190$ & $0.223$ & $0.010$ & $0.043$ & 5 & $4.6$ \\
FreeSolv & high & 33 & $0.43$ & $-0.025$ & $-0.178$ & $0.265$ & $0.112$ & 2 & $1.9$ \\
 & low & 32 & $0.21$ & $-0.171$ & $0.474$ & $0.185$ & $\mathbf{0.830}$ & 2 & $1.5$ \\
Lipophilicity & high & 210 & $0.72$ & $\mathbf{0.092}$ & $\mathbf{0.222}$ & $0.044$ & $\mathbf{0.174}$ & 1 & $2.0$ \\
 & low & 210 & $0.36$ & $\mathbf{0.127}$ & $\mathbf{0.191}$ & $-0.016$ & $0.048$ & 5 & $4.7$ \\
\bottomrule
\end{tabular}
\end{table*}

\begin{table*}[t]
\centering
\footnotesize
\setlength{\tabcolsep}{4pt}
\caption{TabFM on RDKit descriptors compared with RF and TabPFN on the same descriptors, and with tuned Chemprop using the graph alone or the graph with RDKit descriptors. Scores are from the published split. A positive difference means the other model is better. Brackets give the 95\% interval, bold values are significant, and W counts the seeded splits (of 10) that TabFM wins.}
\label{tab:comparisons}
\begin{tabular}{lrrrrrrrr}
\toprule
 & \multicolumn{2}{c}{vs RF} & \multicolumn{2}{c}{vs TabPFN} & \multicolumn{2}{c}{vs Chemprop} & \multicolumn{2}{c}{vs Chemprop + rdkit2d} \\
\cmidrule(lr){2-3} \cmidrule(lr){4-5} \cmidrule(lr){6-7} \cmidrule(lr){8-9}
Dataset & Difference [CI] & W & Difference [CI] & W & Difference [CI] & W & Difference [CI] & W \\
\midrule
ESOL (RMSE) & $-0.051$ [$-0.132$, $0.038$] & 10 & $-0.018$ [$-0.061$, $0.033$] & 6 & $-0.136$ [$-0.250$, $-0.020$] & 10 & $-0.112$ [$-0.204$, $-0.021$] & 7 \\
FreeSolv (RMSE) & $-0.517$ [$-1.051$, $0.095$] & 10 & $-0.028$ [$-0.265$, $0.255$] & 7 & $\mathbf{-0.711}$ [$-1.168$, $-0.239$] & 10 & $\mathbf{-0.795}$ [$-1.453$, $-0.177$] & 10 \\
Lipophilicity (RMSE) & $\mathbf{-0.100}$ [$-0.145$, $-0.053$] & 10 & $0.001$ [$-0.023$, $0.024$] & 8 & $-0.029$ [$-0.091$, $0.031$] & 6 & $\mathbf{-0.081}$ [$-0.142$, $-0.019$] & 8 \\
BBBP (AUC) & $-0.007$ [$-0.035$, $0.021$] & 8 & $0.009$ [$-0.012$, $0.031$] & 5 & $-0.028$ [$-0.067$, $0.011$] & 9 & $-0.011$ [$-0.043$, $0.022$] & 8 \\
BACE (AUC) & $-0.030$ [$-0.060$, $0.000$] & 4 & $0.001$ [$-0.016$, $0.019$] & 5 & $\mathbf{-0.084}$ [$-0.146$, $-0.023$] & 9 & $-0.025$ [$-0.069$, $0.017$] & 10 \\
ClinTox (AUC) & $-0.057$ [$-0.120$, $-0.006$] & 9 & $-0.010$ [$-0.041$, $0.015$] & 9 & $\mathbf{-0.072}$ [$-0.129$, $-0.025$] & 8 & $-0.043$ [$-0.102$, $-0.001$] & 8 \\
SIDER (AUC) & $-0.006$ [$-0.028$, $0.016$] & 7 & $-0.006$ [$-0.020$, $0.008$] & 5 & $\mathbf{-0.080}$ [$-0.119$, $-0.042$] & 10 & $\mathbf{-0.054}$ [$-0.080$, $-0.029$] & 10 \\
\bottomrule
\end{tabular}
\end{table*}
\begin{table*}[t]
\centering
\footnotesize
\setlength{\tabcolsep}{4pt}
\caption{Chemprop against the RF, published split. A positive difference means the RF is better. Brackets give the 95\% interval, bold values are significant, and W counts the seeded splits (of 10) that Chemprop wins.}
\label{tab:chemprop}
\begin{tabular}{lrrrr}
\toprule
 & \multicolumn{2}{c}{vs RF on ECFP4} & \multicolumn{2}{c}{vs RF on rdkit2d} \\
\cmidrule(lr){2-3} \cmidrule(lr){4-5}
Dataset & Difference [CI] & W & Difference [CI] & W \\
\midrule
ESOL (RMSE) & $\mathbf{-0.599}$ [$-0.804$, $-0.381$] & 10 & $0.085$ [$-0.034$, $0.210$] & 3 \\
FreeSolv (RMSE) & $\mathbf{-1.138}$ [$-1.825$, $-0.463$] & 10 & $0.193$ [$-0.160$, $0.640$] & 6 \\
Lipophilicity (RMSE) & $\mathbf{-0.175}$ [$-0.256$, $-0.090$] & 10 & $-0.071$ [$-0.140$, $-0.002$] & 10 \\
BBBP (AUC) & $-0.024$ [$-0.071$, $0.022$] & 5 & $0.021$ [$-0.025$, $0.065$] & 3 \\
BACE (AUC) & $\mathbf{0.121}$ [$0.055$, $0.191$] & 2 & $0.054$ [$-0.008$, $0.117$] & 2 \\
ClinTox (AUC) & $-0.078$ [$-0.303$, $0.064$] & 10 & $0.015$ [$-0.038$, $0.070$] & 7 \\
SIDER (AUC) & $\mathbf{0.053}$ [$0.015$, $0.093$] & 1 & $\mathbf{0.074}$ [$0.035$, $0.113$] & 0 \\
\bottomrule
\end{tabular}
\end{table*}

\begin{table*}[t]
\centering
\footnotesize
\caption{Tabular foundation models against DGT with seed 0, published split, as RMSE (lower is better) or mean ROC-AUC (higher is better). T marks a significant win for the tabular model, and D a significant win for DGT.}
\label{tab:dgt}
\begin{tabular}{lrrrrrrr}
\toprule
Representation & graph & rdkit2d & rdkit2d & rdkit2d & MACCS & MACCS & ECFP4 \\
Dataset & DGT & TabFM & TabPFN & TabFM+ & TabFM & TabPFN & TabPFN \\
\midrule
ESOL (RMSE) & $0.931$ & $0.715$$^{\mathrm{T}}$ & $0.733$$^{\mathrm{T}}$ & $0.711$$^{\mathrm{T}}$ & $1.131$ & $1.143$ & $1.373$$^{\mathrm{D}}$ \\
FreeSolv (RMSE) & $2.666$ & $1.739$$^{\mathrm{T}}$ & $1.767$$^{\mathrm{T}}$ & $1.816$$^{\mathrm{T}}$ & $1.935$$^{\mathrm{T}}$ & $2.009$ & $3.060$ \\
Lipophilicity (RMSE) & $0.669$ & $0.647$ & $0.646$ & $0.641$ & $0.847$$^{\mathrm{D}}$ & $0.831$$^{\mathrm{D}}$ & $0.903$$^{\mathrm{D}}$ \\
BBBP (AUC) & $0.704$ & $0.757$$^{\mathrm{T}}$ & $0.766$$^{\mathrm{T}}$ & $0.752$ & $0.755$ & $0.736$ & $0.711$ \\
BACE (AUC) & $0.847$ & $0.847$ & $0.848$ & $0.852$ & $0.865$ & $0.854$ & $0.874$ \\
ClinTox (AUC) & $0.832$ & $0.947$$^{\mathrm{T}}$ & $0.937$$^{\mathrm{T}}$ & $0.948$$^{\mathrm{T}}$ & $0.914$ & $0.902$ & $0.728$ \\
SIDER (AUC) & $0.627$ & $0.695$$^{\mathrm{T}}$ & $0.689$$^{\mathrm{T}}$ & $0.699$$^{\mathrm{T}}$ & $0.681$$^{\mathrm{T}}$ & $0.661$ & $0.650$ \\
\bottomrule
\end{tabular}
\end{table*}

\begin{table*}[tp]
\centering
\footnotesize
\caption{Run time and peak memory on the published split, with RDKit descriptors (Chemprop: graph), on one machine with an A100 GPU. Times are the median of 2 runs.}
\label{tab:timing}
\setlength{\tabcolsep}{4pt}
\begin{tabular}{llrrrr}
\toprule
Model & Device & Fit (s) & Predict (s) & RAM (GB) & GPU (GB) \\
\midrule
\multicolumn{6}{l}{ESOL (1,128 molecules)} \\
RF & CPU & $5.0$ & $0.12$ & $0.2$ & -- \\
CatBoost & CPU & $26.4$ & $<0.01$ & $0.4$ & -- \\
TabPFN & GPU & $1.4$ & $1.48$ & $2.0$ & $1.8$ \\
TabFM & GPU & $0.6$ & $8.55$ & $13.1$ & $8.9$ \\
Chemprop & GPU & $25.1$ & $11.82$ & $2.6$ & $0.1$ \\
\midrule
\multicolumn{6}{l}{Lipophilicity (4,200 molecules)} \\
RF & CPU & $16.9$ & $0.14$ & $0.4$ & -- \\
CatBoost & CPU & $34.9$ & $<0.01$ & $0.4$ & -- \\
TabPFN & GPU & $1.6$ & $6.97$ & $2.2$ & $3.5$ \\
TabFM & GPU & $1.2$ & $30.61$ & $13.1$ & $17.1$ \\
Chemprop & GPU & $56.1$ & $12.20$ & $2.7$ & $0.2$ \\
\bottomrule
\end{tabular}
\end{table*}